\documentclass[runningheads]{llncs}
\usepackage{graphicx}
\usepackage{caption}
\usepackage{subcaption}
\usepackage{balance}

\usepackage{graphicx}
\usepackage{multirow}
\usepackage{floatrow}
\usepackage{makecell}
\usepackage{booktabs}
\usepackage{wrapfig}
\usepackage[colorlinks,
            linkcolor=blue,
            anchorcolor=blue,
            citecolor=blue,
            urlcolor=blue]{hyperref}
            
\usepackage{pgfplots}
\pgfplotsset{width=6cm,compat=1.9}
\usepackage{amsmath,amssymb}
\usepackage{bm}
\usepackage{arydshln}
\usepackage{booktabs}

\renewcommand{\arraystretch}{0.85}

\usepackage{CJKutf8}

\begin{document}

\title{A Bi-directional Multi-hop Inference Model \\for Joint Dialog Sentiment Classification \\and Act Recognition}
\titlerunning{Bi-directional Multi-hop Inference Model}

\author{Li Zheng\inst{1} \and Fei Li\inst{1}\thanks{Corresponding author.} \and Yuyang Chai\inst{1} \and Chong Teng\inst{1} \and  
Donghong Ji\inst{1} }
\institute{Key Laboratory of Aerospace Information Security and Trusted Computing, Ministry of Education, School of Cyber Science and Engineering, Wuhan University, Wuhan, China \\
\email{\{zhengli,lifei\_csnlp,yychai,tengchong,dhji\}@whu.edu.cn}
}

%\email{{zhengli,lifei_csnlp,yychai,tengchong,dhji}@whu.edu.cn }}
% %
\authorrunning{L. Zheng et al.}

%\title{A Bi-direction Multi-hop Inference Model \thanks{Supported by organization x.}}

\maketitle              % typeset the header of the contribution

\begin{abstract}
The joint task of Dialog Sentiment Classification (DSC) and Act Recognition (DAR) aims to predict the sentiment label and act label for each utterance in a dialog simultaneously. 
However, current methods encode the dialog context in only one direction, which limits their ability to thoroughly comprehend the context. 
Moreover, these methods overlook the explicit correlations between sentiment and act labels, which leads to an insufficient ability to capture rich sentiment and act clues and hinders effective and accurate reasoning. 
To address these issues, we propose a Bi-directional Multi-hop Inference Model (BMIM) that leverages a feature selection network and a bi-directional multi-hop inference network to iteratively extract and integrate rich sentiment and act clues in a bi-directional manner. 
We also employ contrastive learning and dual learning to explicitly model the correlations of sentiment and act labels. 
Our experiments on two widely-used datasets show that BMIM outperforms state-of-the-art baselines by at least 2.6\% on F1 score in DAR and 1.4\% on F1 score in DSC.
Additionally, Our proposed model not only improves the performance but also enhances the interpretability of the joint sentiment and act prediction task.

\end{abstract}

\vspace{-0.8cm}
\section{Introduction}
\vspace{-0.1cm}

\begin{table}[t]
\scriptsize
\centering
%\fontsize{8}{9}\selectfont
\setlength{\tabcolsep}{0.4mm}{
\begin{tabular}{clcc}
\toprule
\textbf{Speaker} &\textbf{Utterances} & \textbf{Act}& \textbf{Sentiment} \\ \midrule
 A & \begin{tabular}[c]{@{}l@{}}$u_1$: There's no way to make a post visible to just your \\ local tl and not federated tl. \end{tabular} 
 & Statement & Negative \\
B & $u_2$: Correct ? & Question & Negative \\
B & $u_3$: I don't think there is. & Answer
 & Negative \\
A &$u_4$: Thanks. & Thanking
 & Positive \\
B &$u_5$: Didn't think so. & Agreement
 & Negative \\
\bottomrule
\end{tabular}}
\caption{A dialog snippet from the Mastodon dataset \cite{DBLP:conf/coling/CerisaraJOL18} for joint dialog sentiment classification and act recognition. 
\vspace{-3mm}}
\label{tab:ex}
\vspace{-8mm}
%\vspace{-10pt}
\end{table}

Dialog Sentiment Classification (DSC) and Act Recognition (DAR) have attracted increasing attention in the field of dialog-based natural language understanding \cite{abs-2306-03975,Feiijcai22DiaRE,Feiijcai22DiaSRL}.
DSC aims to detect the emotion (e.g., negative) expressed by a speaker in each utterance of the dialog, while DAR seeks to assign a semantic label (e.g., question) to each utterance and characterize the speaker's intention.
Recent studies demonstrate that these two tasks are closely relevant, and how to exploit the correlations between them and thoroughly understand the context are key factors.
Thus nichetargeting models are proposed to jointly address these two tasks by utilizing their correlations \cite{DBLP:conf/coling/CerisaraJOL18,DBLP:journals/prl/XuYLLX23,DBLP:conf/acl/XingT22}.

Despite promising performance, most prior approaches only encode the dialog context in one direction, i.e., the chronological order of the utterances~\cite{DBLP:conf/aaai/QinLCNL21,DBLP:journals/prl/KimK18}. 
However, such approaches neglect the subsequent utterances after the target utterance which also play important roles in sentiment classification and act recognition.
As shown in Table~\ref{tab:ex}, the sentiment expressed by $u_2$ ``correct?" is obscure when considering only the dialog context before $u_2$.
Nevertheless, if we check the dialog context after $u_2$, we can observe that the subsequent utterances $u_3$ and $u_5$ from the same speaker both tend to express negative sentiment.
Therefore, it is easy to infer the sentiment label for $u_2$ as ``Negative", 
which highlights the necessity of bi-directional inference.

Moreover, existing works only implicitly exploite the correlations between sentiment and act labels \cite{DBLP:conf/coling/CerisaraJOL18}, or even disregard the correlations at all \cite{DBLP:journals/prl/KimK18}.
The lack of explicit modeling results in an insufficient ability to capture rich sentiment and act clues, which prevents effective and accurate reasoning.
Intuitively, considering the tight associations between sentiments and acts, it is beneficial to explicitly model their correlations.
For instance, the sentiment labels of most utterances in the dialog of Table~\ref{tab:ex} are negative.
A model that solely considers the dialog context is apt to incorrectly predict the sentiment label of $u_4$ as ``Negative''.
In contrast, if the model is capable of explicitly considering the correlations between sentiments and acts, it can deduce that the act label ``Thanking" is more likely to be associated with a positive sentiment label.
Hence, explicitly modeling the correlations is necessary for both interpretability and performance improvement of joint sentiment and act prediction.

In this paper, we propose a Bi-directional Multi-hop Inference Model (BMIM) to model the dialog context in a bi-directional manner and explicitly exploit the correlations of sentiment and act labels.
Firstly, we design a feature selection network to capture sentiment-specific and act-specific features, as well as their interactions, while removing the effect of multi-task confounders.
Next, we leverage a bi-directional multi-hop inference network to iteratively extract and integrate rich sentiment and act clues from front to back and vice versa, emulating human's reasoning and cognition.
Then, we employ contrastive learning and dual learning to explicitly model the correlations of sentiments and acts, meanwhile increasing the interpretability of our model.
Finally, we utilize two classifiers to predict the sentiment and act labels for each utterance based on the refined features from aforementioned modules.

To verify the effectiveness of our model, we conduct experiments on two widely-used datasets for DSC and DAR, namely Mastodon \cite{DBLP:conf/coling/CerisaraJOL18} and Dailydialog \cite{DBLP:conf/ijcnlp/LiSSLCN17}. 
The experimental results show that our model significantly outperforms all state-of-the-art baselines by at least 2.6\% on F1 score in DAR and 1.4\% on F1 score in DSC.
Additionally, we conduct extensive experiments to show that our model has decent interpretability, such as visualization of the correlations between sentiment and act labels, visualization of the correlations between sentiment and act distributions, calculation of casual effect scores \cite{chen-etal-2020-exploring-logically}.
In conclusion, the contributions of this paper can be summarized as follows:
\begin{itemize}
\item We propose a novel bi-directional multi-hop inference model to analyze sentiments and acts in dialogs by understanding dialog contexts based on the way of imitating human reasoning.
\item We employ contrastive learning and dual learning to explicitly model the correlations between sentiments and acts, leading to reasonable interpretability.
\item We conduct extensive experiments on two public benchmark datasets, pushing the state-of-the-art for sentiment and act analyses in dialog. 
% The code will be released to facilitate related research.
\end{itemize}

\vspace{-0.7cm}
\section{Related Work}
\vspace{-0.3cm}
\subsection{Sentiment and Emotion Analyses in NLP}
\vspace{-0.3cm}
Sentiment analysis \cite{Wu0LZLTJ22,shi-etal-2022-effective,FeiCLJZR23,fei-etal-2023-reasoning} has long been an important research topic in natural language processing (NLP), deriving many research directions such as aspect-based sentiment analysis (ABSA) \cite{LIANG2022107643}, emotion detection \cite{FeiZRJ20} and emotion recognition in conversations (ERC) \cite{li-etal-2023-diaasq} and emotion-cause pair extract (ECPE) \cite{DBLP:conf/acl/XiaD19,chen-etal-2022-joint,abs-2306-03969}. 
ABSA focuses on detecting the sentiment polarities of different aspects in the same sentence \cite{Feiijcai22UABSA}.
ECPE considers emotions and their associated causes.
In this paper, we focus on a new scenario, where not only the speaker's sentiments but also their acts should be extracted from the utterances.

\vspace{-0.4cm}
\subsection{Sentiment Classification and Act Recognition}
\vspace{-0.3cm}
Dialog Sentiment Classification and Act Recognition are sentence-level sequence classification problems, and it has been found that they are correlated \cite{DBLP:conf/coling/CerisaraJOL18,DBLP:journals/prl/KimK18}. 
Several joint models are proposed to enhance their mutual interactions, providing a more comprehensive understanding of the speaker's intentions.
Li et al. \cite{DBLP:conf/coling/LiFJ20} propose a context-aware dynamic convolution network (CDCN) to model the local contexts of utterances, but not consider the interactions between tasks.
Qin et al. \cite{DBLP:conf/aaai/QinLCNL21} propose a co-interactive graph attention network (Co-GAT) to consider both contextual information and mutual interaction information, but they ignore the role of label information.
Xing et al. \cite{DBLP:conf/acl/XingT22} propose a dual-task temporal relational recurrent reasoning network (DARER) to achieve prediction-level interactions and estimate label distributions.
However, the aforementioned methods only model the dialog context in a one-way manner, disregarding the explicit correlations of labels between tasks and lacking interpretability.
In contrast, we propose a bi-directional multi-hop inference model to bi-directional capture the dialog context and explicitly model the correlations of sentiment and act labels using contrastive learning and dual learning.

\vspace{-0.4cm}
\subsection{Contrastive Learning and Dual Learning}
\vspace{-0.3cm}
Contrastive learning is a label-efficient representation learning mechanism that enhances the proximity of positive samples and increases the distance between negative samples, which exhibits advantages in various domains \cite{huang-etal-2022-conversation,DBLP:conf/nlpcc/ChaiTFWLCJL22}.
Besides, dual learning has been widely adopted in various tasks \cite{DBLP:conf/acl/SuHC19,DBLP:conf/acl/CaoZLLY19,0001LJL22}, including Question Answering/Generation \cite{FeiMatchStruICML22} and Automatic Speech Recognition/Text-to-Speech \cite{DBLP:conf/acl/CaoZLLY19}.
The primal task and the dual task form a closed loop, generating informative feedback signals that can benefit both tasks.
In this work, we present two approaches to enhance the interpretability of our model: contrastive learning and dual learning. 
Contrastive learning integrates logical dependencies between labels into dialog comprehension, whereas dual learning promotes mutual learning between the two tasks by incorporating logical dependencies.
% into dual task reasoning.

\begin{figure*}[!t]
    % \makebox[\textwidth][c]{\includegraphics[width=0.8\textwidth]{figure//example}}
	\centering
	\includegraphics[scale=0.38]{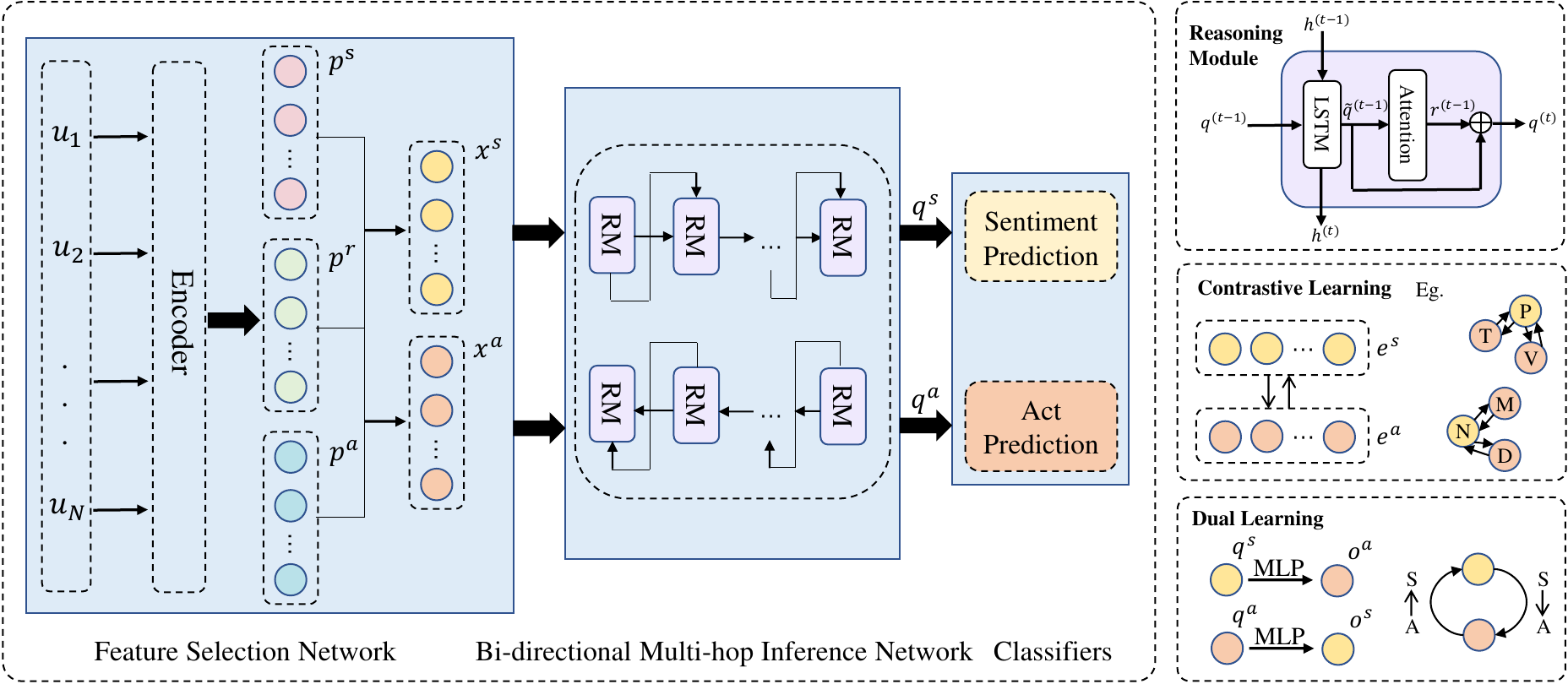}
	\vspace{-0.5\baselineskip}
	\caption{The overview of our model. $p^s$, $p^a$ and $p^r$ denote sentiment, act and shared features. RM means Reasoning module (cf. Section~\ref{sec:bi_infer_net}). Contrastive learning is applied on sentiment and act label embeddings $e^s$ and $e^a$ to pull the ones frequently occurring in the same utterance close.
 P and N represent sentiment labels, positive and negative, while T, V, M and D represent act labels, namely thanking, explicit performative, sympathy and disagreement. 
 When using dual learning, sentiment and act predictions are no longer parallel, they are conducted as pipeline to mimic causal inference.(cf. Section~\ref{sec:cl_dl} for details).
\vspace{-6mm}
 }
	\label{fig:model}
\end{figure*}

\vspace{-0.5cm}
\section{Methodology}
\vspace{-0.3cm}
In this paper, we propose a bi-directional multi-hop inference model to extract features from the dialog context in both directions and explicitly exploit the correlations between sentiment and act labels.
The architecture of our model is illustrated in Figure~\ref{fig:model} and comprises four components. 
First, we design a feature selection network to extract specific sentiment and act features and purify the shared features between them.
Then, we propose a bi-directional multi-hop inference network to encode the dialog context from front to end and vice versa.
Afterward, we explicitly model the correlations between sentiment and act labels employing contrastive learning and dual learning, respectively.
Finally, we utilize two classifiers for dialog sentiment and act prediction.

\vspace{-0.3cm}
\subsection{Task Definition}
\vspace{-0.2cm}
Let $U=\{u_1, u_2, ..., u_N\}$ be a dialog, where $N$ is the number of utterances. 
Our goal is to map utterance sequences $(u_1, u_2, ... , u_N)$ to the corresponding utterance sequence sentiment  labels $(y_{1}^s, y_{2}^s, ..., y_{N}^s)$ and act labels $(y_{1}^a, y_{2}^a, ..., y_N^a)$.

\vspace{-0.3cm}
\subsection{Utterance Encoding}
\vspace{-0.2cm}

Following Xing et al. \cite{DBLP:conf/acl/XingT22}, we also apply BiLSTM \cite{DBLP:journals/neco/HochreiterS97} as the encoder to yield initial utterance representations $U = \{u_1,u_2, …, u_N\}$.
Next, we leverage a feature selection network (FSN) \cite{DBLP:conf/emnlp/YanZFZW21} to extract task-specific and interactive features and remove the influence of redundant information.
FSN divides neurons into three partitions (sentiment, act, and shared) at each time step, generating task-specific features by selecting and combining these partitions and filtering out irrelevant information. 
Specifically, at the i-th time step, we generate two task-related gates:
\setlength\abovedisplayskip{3pt}
\setlength\belowdisplayskip{3pt}
\begin{equation}
\label{eq:1}
\begin{split}
s_i = Cummax(Linear([u_i;h_{i-1}])), 
 a_i = 1-Cummax(Linear([u_i;h_{i-1}]))
\end{split}
\end{equation}
where $Cummax(\cdot)$ denotes the cumulative maximum operation that performs as a binary gate, $Linear(\cdot)$ refers to a linear transformation, and $h_{i-1}$ represents the hidden state of the $(i-1)$-th utterance.
Each gate corresponds to a specific task and divides the utterance representations into two segments according to their relevance to the respective task. 
With the joint efforts of the two gates $s_i$ and $a_i$, the utterance representations can be divided into three partitions: the sentiment partition $p^s_i$ , the act partition $p^a_i$ and the shared partition $p^r_i$.
Next, we concatenate the sentiment and act partition separately with the shared partition to gain task-specific sentiment and act feature representations $x^s_i$ and $x^a_i$:
\setlength\abovedisplayskip{3pt}
\setlength\belowdisplayskip{3pt}
\begin{equation}
\label{eq:2}
\begin{split}
x^s_i = tanh(p^s_i) + tanh(p^r_i),
\quad x^a_i = tanh(p^a_i) + tanh(p^r_i)
\end{split}
\end{equation}

\vspace{-0.4cm}
\subsection{Bi-directional Multi-hop Inference Network}
\label{sec:bi_infer_net}
\vspace{-0.2cm}

To imitate human reasoning and mine the internal dependencies of utterances to thoroughly understand the context, we propose a bi-directional multi-hop inference network.
Concretely, in the t-th turn, we adopt the LSTM network to learn intrinsic logical order and integrate contextual clues in the working memory, formulated as:
\setlength\abovedisplayskip{3pt}
\setlength\belowdisplayskip{3pt}
\begin{equation}
\label{eq:3}
q_i^{(0)} = W_qx_i + b_q 
\end{equation}
\begin{equation}
\label{eq:4}
\tilde{q}_i^{(t-1)},h_i^{(t)} = \overrightarrow{LSTM}(q_i^{(t-1)},h_i^{(t-1)})
\end{equation}
where $W_q$ and $b_q$ are learnable parameters. $x_i$ can be either $x^s_i$ or $x^a_i$.
$h_i^{(t)}$ is the working memory, which stores and updates the previous memory $h_i^{(t-1)}$ and guides the next turn of clue extraction. $t$ denotes the number of inference steps.

In order to mimic human retrieval and reasoning processes, we utilize an attention mechanism to mine relevant contextual clues:
\setlength\abovedisplayskip{1pt}
\setlength\belowdisplayskip{1pt}
\begin{equation}\small
e_{ij}^{(t-1)} = f(x_j,\tilde{q}^{(t-1)}_i), \, \alpha_{ij}^{(t-1)} = \frac{exp(e^{(t-1)}_{ij})}{\sum_{j=1}^{N}exp(e^{(t-1)}_{ij})}  
, \, r^{(t-1)}_i = \sum_{j=1}^{N}\alpha_{ij}^{(t-1)}x_j 
\end{equation}
% \begin{equation}\small
% \alpha_{ij}^{(t-1)} = \frac{exp(e^{(t-1)}_{ij})}{\sum_{j=1}^{N}exp(e^{(t-1)}_{ij})}  
% \end{equation}
% \begin{equation}\small
% r^{(t-1)}_i = \sum_{j=1}^{N}\alpha_{ij}^{(t-1)}x_j 
% \end{equation}
where $f$ is a dot product function.
Then, we concatenate the output of inference process $\tilde{q}^{(t-1)}_i$ with the resulting attention readout $r^{(t-1)}_i$ to form the next-turn queries $q^{(t)}_{f\to b}$ and $q^{(t)}_{b\to f}$ from front to back and vice versa to explore contextual dependencies:
\setlength\abovedisplayskip{3pt}
\setlength\belowdisplayskip{3pt}
\begin{equation}
\begin{split}
q^{(t)}_{f\to b} = [\tilde{q}^{(t-1)}_{f\to b};r^{(t-1)}_{f\to b}],
\quad q^{(t)}_{b\to f} = [\tilde{q}^{(t-1)}_{b\to f};r^{(t-1)}_{b\to f}]
\end{split}
\end{equation}

Based on the above output vectors, the final sentiment and act representations $q^s$ and $q^a$ with rich contextual clues can be defined as a concatenation of both vectors:
\setlength\abovedisplayskip{3pt}
\setlength\belowdisplayskip{3pt}
\begin{equation}
\begin{split}
q^s = [q^s_{f\to b};q^s_{b\to f}],
\quad q^a = [q^a_{f\to b};q^a_{b\to f}]
\end{split}
\end{equation}

\vspace{-0.4cm}
\subsection{Contrastive Learning and Dual Learning}
\label{sec:cl_dl}
\vspace{-0.25cm}

\textbf{Contrastive Learning.} 
To model the correlations between labels, bringing related sentiment and action labels closer while pushing unrelated labels away, we employ contrastive learning
\cite{DBLP:conf/nlpcc/ChaiTFWLCJL22}.
Our contrastive loss function is defined as:
\setlength\abovedisplayskip{1pt}
\setlength\belowdisplayskip{1pt}
\begin{equation}
    \mathcal{L} ^{cl}=\sum_{i=1}^{l} \frac{-1}{\left | \mathcal{P}  \right | } \sum_{p\in\mathcal{P} }^{}\log_{}{\frac{\exp (e_i\cdot e_p/\tau  )}{ {\textstyle \sum_{p\in\mathcal{P}}\exp(e_i\cdot e_p/\tau  )+  \sum_{n \in \mathcal{N} }\exp (e_i\cdot e_n / \tau ) + \varepsilon  } } } 
\end{equation}
where 
$l$ is the total category size of sentiment and act labels and $e_i$ is the label embedding representation.
The positive set $\mathcal{P}$ contains the indexes of co-existed labels with the label $e_i$ in the training batch, while the negative set $\mathcal{N}$ contains the indexes of labels that never co-existed with $e_i$ in the training set.

\begin{figure*}[!t]
    % \makebox[\textwidth][c]{\includegraphics[width=0.8\textwidth]{figure//example}}
	\centering
	\includegraphics[scale=0.5]{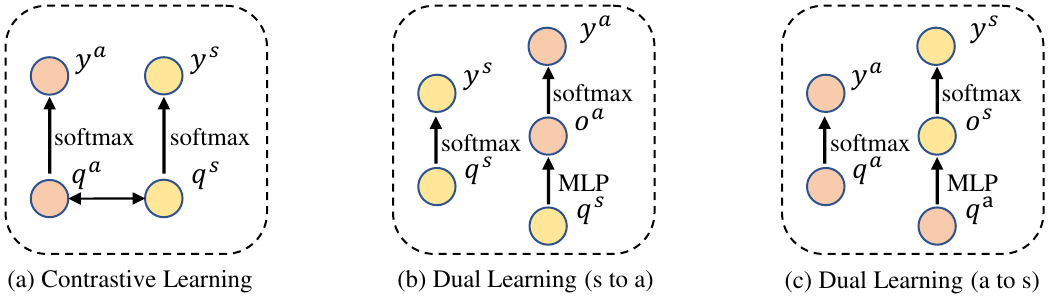}
	\vspace{-0.5\baselineskip}
	\caption{Three architectures of the output layer of our model. Architecture (a) applies contrastive learning on sentiment and act representations, so sentiment and act distributions can be predicted in parallel. Architecture (b) and (c) apply dual learning where act or sentiment distributions are predicted grounded on the features given by the sentiment or act task.
 \vspace{-5mm}
}
	\label{fig:three}
\end{figure*}

\noindent \textbf{Dual Learning.} 
We incorporate dual learning into our approach to facilitate mutual learning of the two tasks, allowing for a more comprehensive understanding and utilization of the logical dependencies between sentiment and act labels. 
Rather than solely considering inter-task duality, we also take into account the causalities of sentiments on acts and acts on sentiments to further improve the model performance.
Specifically, we consider the impact of sentiment on act and predict the act score $o^{a}= MLP(q^{s})$, based on a multi-layer perceptron and sentiment representation $q^{s}$.
Then we employ a classifier to predict dialog act and sentiment for each utterance as $ y^a = Softmax(o^{a}) = P(a|s;\theta _{s \to a})$.
Similarly, we can perform such process reversely and model the impact of act on sentiment, formulated as: $y_{}^s = Softmax(MLP(q^{s}))$.
To enforce the dual learning process, we have the following loss:
\setlength\abovedisplayskip{3pt}
\setlength\belowdisplayskip{3pt}
\begin{equation}
    \mathcal{L}_{} ^{dl} = (log\hat{P}(s) + logP(a|s;\theta _{s \to a})-log\hat{P}(a) -logP(s|a;\theta _{a \to s}))^2
\end{equation}
where $y_{}^a=P(a|s;\theta _{s \to a})$ and $y_{}^s=P(s|a;\theta _{a \to s})$.
Note that the true marginal distribution of data $P(s)$ and $P(a)$ are often intractable, so here we substitute them with the approximated empirical marginal distribution $\hat{P}(a)$ and $\hat{P}(s)$ \cite{DBLP:conf/acl/SuHC19}.

\vspace{-0.4cm}
\subsection{Joint Training}
\vspace{-0.2cm}
The training loss is defined as the cross-entropy loss between predicted label distributions and ground-truth label distributions in the training set:
\setlength\abovedisplayskip{1pt}
\setlength\belowdisplayskip{1pt}
\begin{equation}
    \mathcal{L}^s = -\sum_{i = 1}^{N} \sum_{j = 1}^{N_s} \hat{y}_{ij}^slog(y_{ij}^s), 
    \quad \mathcal{L}^a = -\sum_{i = 1}^{N} \sum_{j = 1}^{N_a} \hat{y}_{ij}^alog(y_{ij}^a) 
\end{equation}
where ${y}_{ij}^s$, ${y}_{ij}^a$, $\hat{y}_{ij}^s$ and $\hat{y}_{ij}^a$ are the predicted and gold sentiment and act distributions for the i-th utterance.
$N_s$ and $N_a$ are the numbers of sentiment and act labels. 
The aforementioned 4 losses can be combined and applied on the three architectures in Figure~\ref{fig:three}, where Equation~\ref{eq:13}, \ref{eq:14}, \ref{eq:15} corresponding to Architecture (a), (b) and (c) respectively:
\setlength\abovedisplayskip{1pt}
\setlength\belowdisplayskip{1pt}
\begin{equation}
\label{eq:13}
    \mathcal{L} ^c = \mathcal{L}^s  + \mathcal{L}^a + \mathcal{L} ^{cl} 
\end{equation}
\begin{equation}
\label{eq:14}
    \mathcal{L} ^{d,s \to a} = \mathcal{L}^s  + \mathcal{L}^a + \mathcal{L}_{s \to a} ^{dl}
\end{equation}
\begin{equation}
\label{eq:15}
    \mathcal{L} ^{d,a \to s} = \mathcal{L}^s  + \mathcal{L}^a + \mathcal{L}_{a \to s} ^{dl}
\end{equation}

\vspace{-0.5cm}
\section{Experiments}
\vspace{-0.3cm}
\subsection{Datasets and Evaluation Metrics}
\vspace{-0.2cm}

\textbf{Datasets.} We assess the efficacy of our model on two publicly dialog datasets, Mastodon \cite{DBLP:conf/coling/CerisaraJOL18} and Dailydialog \cite{DBLP:conf/ijcnlp/LiSSLCN17}.
The Mastodon dataset consists 269 dialogs with 1075 utterances allocated to training and 266 dialogs with 1075 utterances reserved for testing.
It encompasses 3 sentiment categories and 15 act categories.
To ensure consistency with Cerisara et al. \cite{DBLP:conf/coling/CerisaraJOL18}, we follow the same partition scheme.
% The Dailydialog dataset includes 11,118 dialogs for training, 1,000 for validation and 1,000 for testing, encompassing 7 sentiment categories and 4 act categories.

\noindent \textbf{Evaluation Metrics.} Following previous works \cite{DBLP:conf/coling/CerisaraJOL18,DBLP:conf/acl/XingT22}, we exclude the neutral sentiment label in DSC.
For DAR, we employ the average of the F1 scores weighted by the prevalence of each dialog act on Mastodon. 
While on DailyDialog, we adopt the macro-averaged Precision (P), Recall (R) and F1 as the major metrics to measure the effectiveness of our model in both tasks.

\vspace{-0.2cm}
\subsection{Baseline Systems}
\vspace{-0.2cm}
To verify the effectiveness of the BMIM, we compare it with the following state-of-the-art baselines, which are categorized into three groups based on their modeling approaches. 
The first group solely focuses on modeling the context or interaction information, which includes JoinDAS \cite{DBLP:conf/coling/CerisaraJOL18} and IIIM \cite{DBLP:journals/prl/KimK18}.
The second group considers both the dialog context and implicit interaction information between tasks, and includes DCR-Net \cite{DBLP:conf/aaai/QinCLN020}, BCDCN \cite{DBLP:conf/coling/LiFJ20} , Co-GAT \cite{DBLP:conf/aaai/QinLCNL21} and TSCL \cite{DBLP:journals/prl/XuYLLX23}.
The third group utilizes implicit label information and includes DARER \cite{DBLP:conf/acl/XingT22}.
Notably, all baselines only employ unidirectional modeling of the dialog context.

%\vspace{-0.2cm}
\subsection{Overall Results}
%\vspace{-0.2cm}

\begin{table*}[t]
\centering
\scriptsize

\setlength{\tabcolsep}{1mm}{
\scalebox{0.9}{
\begin{tabular}{ccccccccccccc}
\toprule
\multirow{4}{*}{Models} & \multicolumn{6}{c}{Mastodon}                  & \multicolumn{6}{c}{DailyDialog}   \\ 

\cmidrule(lr){2-7} \cmidrule(lr){8-13}
            & \multicolumn{3}{c}{DSC} & \multicolumn{3}{c}{DAR} & \multicolumn{3}{c}{DSC} & \multicolumn{3}{c}{DAR}\\ \cmidrule(lr){2-4}\cmidrule(lr){5-7}\cmidrule(lr){8-10} \cmidrule(lr){11-13}
                        & P     & R & F1& P & R & F1& P  & R  & F1& P & R & F1\\ \midrule
    JointDAS            &36.1   & 41.6  & 37.6  & 55.6  & 51.9  & 53.2  & 35.4  & 28.8  & 31.2  & 76.2  & 74.5  & 75.1  \\ 
     IIIM    & 38.7  & 40.1  & 39.4  & 56.3  & 52.2  & 54.3  & 38.9  & 28.5  & 33.0  & 76.5  & 74.9  & 75.7  \\  \midrule
  DCR-Net& 43.2  & 47.3  & 45.1  & 60.3  & 56.9  & 58.6  & 56.0  & 40.1  & 45.4  & 79.1  & 79.0  & 79.1  \\  
  BCDCN &38.2 &62.0 &45.9 &57.3 &61.7 &59.4 &55.2 &45.7 &48.6 &80.0 &80.6 &80.3\\ 
   Co-GAT & 44.0  & 53.2  & 48.1  & 60.4  & 60.6  & 60.5  & 65.9  & 45.3  & 51.0  & 81.0  & 78.1  & 79.4  \\ 
  TSCL & 46.1 & 58.7 & 51.6 & 61.2 & 61.6& 60.8 & 56.6 & 49.2 & 51.9 & 78.8 & 79.8 & 79.3 \\   
  DARER & 56.0  & 63.3  & 59.6  & 65.1  & 61.9  & 63.4  & 60.0  & 49.5  & 53.4  & 81.4  & 80.8  & 81.1  \\ 
  \midrule
   BMIM$^{c}$ & 58.2 & 64.0 & \textbf{61.0} & 68.2 & 62.7 & \textbf{65.3} & 60.0 & 49.7 & 54.3 & 83.9 & 83.2 & 83.5 \\
   BMIM$^{d,s \to a}$ & 58.0 & 62.2 & 60.0 & 67.2 & 61.8 & 64.4 & 59.9 & 49.6 & 54.3 & 83.9 & 83.2 & \textbf{83.7} \\
BMIM$^{d,a \to s}$ & 58.7 & 62.4 & 60.5 & 67.1 & 61.8 & 64.3 & 60.0 & 49.7 & \textbf{54.4} & 83.8 & 83.1 & 83.5 \\

 \bottomrule
\end{tabular}}
}
\caption{Comparison of our model with baselines on Mastodon and Dailydialog datasets. 
BMIM$^c$, BMIM$^{d,s \to a}$ and BMIM$^{d,a \to s}$correspond to architecture (a), (b) and (c) in Figure~\ref{fig:three}, respectively.
\vspace{-3mm}}
\vspace{-10pt}
\label{table:results}
\vspace{-3mm}
\end{table*}

The experimental results for both DSC and DAR tasks are presented in Table~\ref{table:results}.
Our method demonstrates obvious advantages over other  state-of-the-art baselines for both tasks. 
For instance, on Mastodon, our model achieves the best performance when employing contrastive learning, surpassing the best baseline (DARER) with an absolute improvement of 1.4\% F1 score on the DSC task and 1.9\% F1 score on the DAR task.
Similarly, on DailyDialog, dual learning with the sentiment-to-act approach works best for the DAR task, outperforming the best baseline (DARER) by 2.6\% F1 score. 
When using dual learning with the act-to-sentiment approach, the performance on DSC tasks is the best, with a 1.0\% higher F1 score than the best baseline (DARER).
We attribute the performance improvement to three aspects:
(1) Our framework implements bidirectional modeling of dialog contexts, which infers context from a perspective that aligns with human thinking and obtains contextual representations with richer sentiment and act clues.
(2) We design contrastive learning and dual learning to explicitly capture the correlations between sentiment and act labels, effectively utilizing label information.
(3) Our feature selection network filters out task-independent information and optimally applies task-related one to promote performance.

\begin{wraptable}[6]{r}{6cm}
\scriptsize
\vspace{-1.7cm}
\floatsetup{floatrowsep=qquad,captionskip=3pt} \tabcolsep=3.5pt 
\begin{floatrow}
\centering
\label{table:ablation}
\renewcommand\arraystretch{1}
\ttabbox{\caption{Results of the ablation study on F1 scores.}\label{table:ablation}}{%
\begin{tabular}{ccccc}
\toprule
\multirow{2}{*}{Variants} & \multicolumn{2}{c}{Mastodon} & \multicolumn{2}{c}{DailyDialog} \\ \cmidrule(lr){2-3} \cmidrule(lr){4-5}
                        & DSC            & DAR            & DSC              & DAR             \\ \midrule
        BMIM$^c$         & \textbf{61.0}       &   \textbf{65.3}       &  \textbf{54.3}        &  \textbf{83.5}        \\ \hline
         w/o BMIN        & 58.2        &  63.6     &  51.8        &  81.7        \\ 
  w/o CL or DL   & 60.2        &   64.3      & 53.6         &  82.7              \\ 
  w/o FSN                & 59.5         & 64.1      &  52.9        &  82.5       \\

  \bottomrule
\end{tabular}}
\end{floatrow}
\end{wraptable}

\vspace{-0.2cm}
\subsection{Ablation Study}
\vspace{-0.2cm}

We conduct ablation experiments to assess the contribution of each component in our model.
As depicted in Table \ref{table:ablation}, we observe that no variants can compete with the complete model, implying the indispensability of each component for the task.
Specifically, the F1 score decreases most heavily without the bi-directional multi-hop inference network, which indicates that it has a significant effect on modeling the complex structural information of dialog.
Furthermore, we investigate the necessity and effectiveness of contrastive learning and dual learning by removing these two modules. 
The sharp drops of results demonstrate that either contrastive learning or dual learning plays an important role in capturing explicit correlations between labels.
Besides, removing the feature selection nerwork results in a distinct performance decline.
This finding implies that efficient utilization of task-related information is able to enhance our model performance.

\vspace{-0.3cm}
\subsection{Visualizing the Effectiveness of Contrastive Learning} 
\vspace{-0.2cm}

\begin{figure*}[!t]
	\centering
	\includegraphics[scale=0.5]{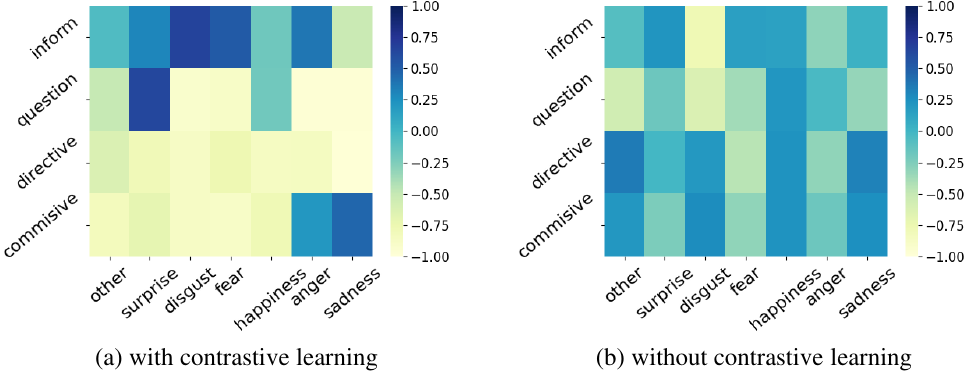}
	\vspace{-0.5\baselineskip}
	\caption{Visualization of the correlations between sentiment and act labels. 
 \vspace{-7mm}}
\vspace{-4mm}
	\label{fig:contra}
\end{figure*}

We perform a visualization study in Figure~\ref{fig:contra} to investigate the effectiveness of contrastive learning and demonstrate that it assists our model in learning the correlations between sentiment and act labels.
Figure~\ref{fig:contra} (a) illustrates that when using contrastive learning, each type of sentiment label has strongly and weakly correlated act labels.
Strongly correlated labels are placed closer together, while weakly correlated labels are farther apart. 
In contrast, as depicted in Figure~\ref{fig:contra} (b), without contrastive learning, the correlations between labels is not significant, which limites the ability to effectively utilize the information between labels for modeling purposes.
These observations reveal the explicit correlations between sentiment and act labels and enhance the interpretability of our model. 
It also proves that our contrastive learning can proficiently model and utilize the correlations between labels to facilitate model inference.

\vspace{-0.3cm}
\subsection{Visualizing the Interpretability of Dual Learning}
\vspace{-0.2cm}

\begin{figure*}[!t]
	\centering
	\includegraphics[scale=0.58]{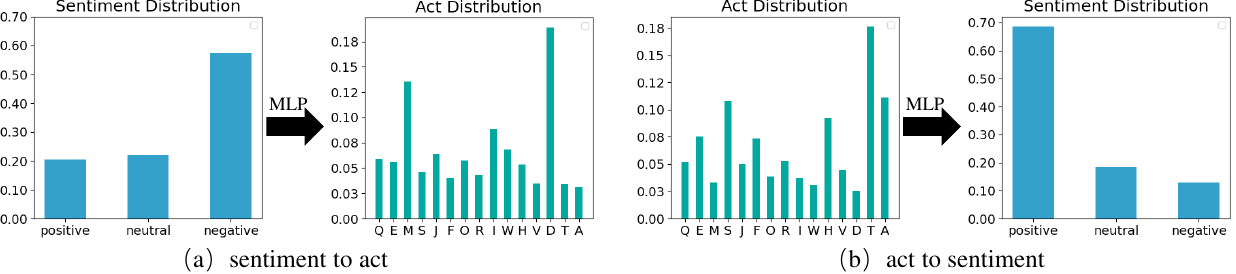}
	\vspace{-0.5\baselineskip}
	\caption{Visualization of the correlations between sentiment and act distributions.
\vspace{-10mm}}
\vspace{-5mm}
	\label{fig:dual}
\end{figure*}

With the attempt to better understand how dual learning exploits explicit correlations between labels, we select two examples corresponding to the architecture (b) and (c) in Figure~\ref{fig:three}, and visualize the distributions of sentiment and act labels in these examples.
The results in Figure~\ref{fig:dual} indicate that when the distribution value of the ``negative" sentiment label reaches its maximum, the model is more likely to predict act labels such as ``disagreement" (D) or ``symmetry" (M). 
Similarly, larger distributions of act labels such as ``thinking" (T), ``agreement" (A), and ``suggestion" (S) make it easier for the model to predict a ``positive" sentiment label. 
In summary, dual learning effectively utilizes explicit correlations between labels, enhancing the model performance. 
This approach also improves the interpretability of such correlations, which aligns with human cognition.

\vspace{-0.3cm}
\subsection{Evaluating Causal Effect Scores of Our Model}
\vspace{-0.2cm}

\begin{wraptable}[11]{R}{5.3cm}
%\begin{wraptable}[6]{r}{6cm}
\scriptsize
\vspace{-0.8cm}
\floatsetup{floatrowsep=qquad,captionskip=5pt} \tabcolsep=2pt 
\begin{floatrow}
\centering
\renewcommand\arraystretch{1}
\ttabbox{\caption{The estimated causal effects of sentiment and act labels on Mastodon. 
A higher score indicates more causal relationship.
``-" denotes that the estimated effect is below 0.001.}\label{tab:causal}}{%
\begin{tabular}{lccc}
\hline

\hline
        & Positive     & Neutral   & Negative \\ \hline
 Thanking   & 0.32 & - & - \\  
    Greeting & 0.28 & - & - \\ 
    Sympathy & - & 0.14 & 0.26 \\
    Agreement & 0.34 & - & 0.32 \\
    Disagreement & - & 0.07 & 0.34 \\
    \hline

\end{tabular}}
%\end{table}%
\end{floatrow}
\end{wraptable}

To establish causal relationships between sentiment and act labels, which aligns with human prior knowledge, we estimate causal effects \cite{chen-etal-2020-exploring-logically} between labels and present some results in Table~\ref{tab:causal}.
The act labels ``Thanking" and ``Greeting" have causal effects on ``Positive" sentiment with scores of 0.32 and 0.28, respectively, and no impact on other sentiment types. 
In addition, the causal relationship between sentiment and act labels is not one-to-one.
Rows 4-6 of Table~\ref{tab:causal} reveal that all act labels have significant causal relationships with two sentiment labels. 
Our quantitative analysis of the causal relationship between labels enhances the interpretability of our model and emphasizes the significance of explicitly modeling label information to improve performance.

\vspace{-0.2cm}
\subsection{Performances Using Different Pre-trained Language Models} 
\vspace{-0.2cm}

\begin{table}[!htbp]
\scriptsize
\floatsetup{floatrowsep=qquad,captionskip=9pt} 
  \centering
\caption{Results based on different pre-trained language models. \vspace{-3mm}}
\label{table:ptlm}
\setlength{\tabcolsep}{2mm}{
\scalebox{1}{
    \begin{tabular}{lcccccc}
    \toprule
    \multicolumn{1}{c}{\multirow{3}{*}{\quad Models}} & \multicolumn{3}{c}{DSC} & \multicolumn{3}{c}{DAR} \\
\cmidrule(lr){2-4} \cmidrule(lr){5-7}          & P(\%)     & R(\%) & F1(\%)    & P(\%)     & R(\%) & F1(\%) \\
    \midrule
BERT + Linear         & 64.6 & 66.5 & 65.5 & 72.5 & 70.6 & 71.6 \\
BERT + DARER           & 65.5 & 67.3 & 66.4 & 73.1 & 71.3 & 72.2 \\
BERT + BMIM         & {67.9} & {70.3} & \textbf{69.0} & {76.6} & {73.9} & \textbf{75.2} \\ \midrule
RoBERTa + Linear     & 60.0 & 64.6 & 62.2 & 69.7 & 67.0 & 68.4 \\
RoBERTa + DARER        & 60.7 & 65.3 & 62.9  & 70.0 & 67.9 & 68.9 \\
RoBERTa + BMIM      & {62.8} & {67.7} & \textbf{65.1} & {74.2} & {71.4} & \textbf{72.7} \\ \midrule
XLNet + Linear   & 64.9 & 66.4 & 65.6 & 70.8 & 69.1 & 69.9 \\
XLNet + DARER           & 67.3 & 68.4 & 67.8 & 71.9 & 69.5 & 70.7 \\
XLNet + BMIM         & 68.9 & 70.3 & \textbf{69.6} & 74.1 & 73.1 & \textbf{73.6} \\ 
    \bottomrule
    \end{tabular}%
    }
    }
  \label{tab:addlabel}%
  \vspace{-5mm}
\end{table}%
Following Xing et al. \cite{DBLP:conf/acl/XingT22}, we also explore three pre-trained models, BERT, Ro-BERTa and XLNet in our framework. 
In this study, we replace the BiLSTM utterance encoder with these pre-trained models, while retaining the other components.
We compare our approach with DARER, a competitive baseline, under different encoders and presented the results in Tabel~\ref{table:ptlm}.
A single pre-trained encoder yields promising results, which highlights the excellent language comprehension abilities of pre-trained models.
Moreover, our model consistently outperforms DARER on both tasks, irrespective of whether pre-trained models are used or not. 
Notably, our model wins DARER over 3.0\% F1 score on DAR task when using BERT.
Additionally, the performance gaps are further enlarged when using only a linear classifier on the pre-trained model.
These outcomes indicate that our method can well model the dialog context bidirectionally while also exploiting the explicit correlations between labels.

\vspace{-4mm}
\section{Conclusion}
\vspace{-3mm}
In this work, we propose a Bi-directional Multi-hop Inference Model to tackle the joint task of DSC and DAR. 
BMIM leverages a feature selection network, a bi-directional multi-hop inference network, as well as contrastive learning and dual learning to iteratively extract and integrate abundant sentiment and act clues in a bi-directional manner, explicitly model the correlations of sentiment and act labels.
Experimental results on two datasets demonstrate the effectiveness of our model, achieving state-of-the-art performance. 
Extensive analysis further confirms that our approach can proficiently comprehend the dialog context in both directions, and exploit the correlations between sentiment and act labels for better performance and interpretability.
\vspace{-3mm}

\subsubsection{Acknowledgment.}
This work is supported by the National Key Research and Development Program of China (No. 2022YFB3103602) and the National Natural Science Foundation of China (No. 62176187).

% \begin{theorem}
% This is a sample theorem. The run-in heading is set in bold, while
% the following text appears in italics. Definitions, lemmas,
% propositions, and corollaries are styled the same way.
% \end{theorem}
%
% the environments 'definition', 'lemma', 'proposition', 'corollary',
% 'remark', and 'example' are defined in the LLNCS documentclass as well.
%
% \begin{proof}
% Proofs, examples, and remarks have the initial word in italics,
% while the following text appears in normal font.
% \end{proof}
% For citations of references, we prefer the use of square brackets
% and consecutive numbers. Citations using labels or the author/year
% convention are also acceptable. The following bibliography provides
% a sample reference list with entries for journal
% articles~\cite{ref_article1}, an LNCS chapter~\cite{ref_lncs1}, a
% book~\cite{ref_book1}, proceedings without editors~\cite{ref_proc1},
% and a homepage~\cite{ref_url1}. Multiple citations are grouped
% \cite{ref_article1,ref_lncs1,ref_book1},
% \cite{ref_article1,ref_book1,ref_proc1,ref_url1}.
\bibliographystyle{splncs04}
%\bibliography{mybibliography}
%

\bibliography{ref}





% LNCS Homepage, \url{http://www.springer.com/lncs}. Last accessed 4
%\end{thebibliography}

\end{document}